\documentclass[fleqn,10pt,twocolumn]{AROB}
\usepackage{subcaption}
\usepackage[hyphens]{url}

\title{Reinforcement Learning with a Focus on Adjusting Policies to Reach Targets}

\author{Akane Tsuboya${}^{1\dagger}$, Yu Kono${}^{1}$, Tatsuji Takahashi${}^{1}$}
\speaker{Akane Tsuboya}

\affils{${}^{1}$Tokyo Denki University, Saitama, Japan\\
(Tel: +81-49-296-1642; E-mail: tatsujit@mail.dendai.ac.jp)\\
}
\abstract{
}

\abstract{
The objective of a reinforcement learning agent is to discover better actions through exploration.
However, typical exploration techniques aim to maximize rewards, often incurring high costs in both exploration and learning processes.
We propose a novel deep reinforcement learning method, which prioritizes achieving an aspiration level over maximizing expected return.
This method flexibly adjusts the degree of exploration based on the proportion of target achievement.
Through experiments on a motion control task and a navigation task, this method achieved returns equal to or greater than other standard methods.
The results of the analysis showed two things: our method flexibly adjusts the exploration scope, and it has the potential to enable the agent to adapt to non-stationary environments.
These findings indicated that this method may have effectiveness in improving exploration efficiency in practical applications of reinforcement learning.
}

\begin{document}

\maketitle
\section{Introduction}
Reinforcement learning is a field of machine learning that has applications in a wide range of domains, including playing Go \cite{Silver17}, solving combinatorial optimization problem \cite{Kool19}, and controlling fusion reactor \cite{Degrave22}.
In this field, agents learn appropriate policies to maximize return through interacting with the environment.
The agent must balance between ``exploration,'' where it seeks new experiences, and ``exploitation,'' where it utilizes past experiences.
The less exploration is exhaustive, the more difficult it becomes for the agent to find appropriate policies.
The simplest exploration method is random selection, such as $\epsilon$-greedy policy.
By just using this policy, the Deep Q-Network (DQN) achieved performance comparable to or exceeding human gameplay in about half of the Atari games it played \cite{Mnih15}.
If designers ensure the resources for learning, such as sufficient time and computational power, the agent can exhaustively explore the environment.
In such cases, random exploration is effective.

In practical applications, however, quick and efficient achievement of a specific level of return often comes to priority rather than exhaustive exploration.
For example, in robotic control, failures in exploration pose a risk of damaging expensive hardware.
In power control, the important thing is to quickly secure an electric supply above a certain level.
Random exploration has the possibility of visiting all states, but it is an unsuitable for quickly and efficiently discovering appropriate policies.
In this respect, expanding further the scope of reinforcement learning requires more efficient exploration methods \cite{Schwarzer23}.

We focused on target-oriented exploration to enhance exploration efficiency.
Previously, we proposed a method called Risk-sensitive Satisficing (RS), which prioritizes achieving an aspiration level over maximizing expected return \cite{Takahashi16}.
Past studies demonstrated the superior performance of RS in bandit problems and reinforcement learning tasks.
In particular, it provides performance guarantees in the Bernoulli bandit problem \cite{Tamatsukuri19}.
This capability is especially useful in deep reinforcement learning, where trial and error tends to be costly.

In this study, we propose Regional Stochastic Risk-sensitive Satisficing ($\mathrm{RS}^2$), an extension of RS for deep reinforcement learning.
The $\mathrm{RS}^2$ agent explores until it finds a particular action whose value exceeds the aspiration level and then continues to exploit the action.
If the value of the action falls below the aspiration level, the agent returns to exploration.
$\mathrm{RS}^2$ represents the ``reliability'' of an action as the frequency at which the agent has selected the action.
This frequency provides additional information for prioritizing actions during exploration.
$\mathrm{RS}^2$ determines its exploration distribution based on this reliability.
We used a dynamic meta-mechanism that adjusts the aspiration level for each state based on the difference between actual and ideal returns.
This mechanism supports flexible adjustment of the degree of exploration.

We tested our method on both a motion control task and a navigation task.
In the experiments, we introduced $\mathrm{RS}^2$ as a behavior policy for DQN, a standard deep reinforcement learning algorithm.
First, we tested whether it has the ability to achieve high returns in these two types of tasks.
Second, we analyzed its exploration behaviors in the navigation task.

\section{Related Work}
Our method, that is an exploration policy for deep reinforcement learning, is an extension of RS.
First, this section introduces studies aimed at improving exploration efficiency in deep reinforcement learning.
Second, we describe studies on target-oriented reinforcement learning and RS.

\subsection{Enhancing exploration efficiency in reinforcement learning}
Many studies have been conducted on methods of exploration in environments with sparse rewards.
When rewards are sparse, it is difficult for the agent to learn the action-reward mapping.
A common solution is to utilize intrinsic rewards from within the agent.
Burda et al. (2018) designed intrinsic rewards based on the error between a target network that was initialized randomly and a predictor network that mimics its outputs \cite{Burda18}.
This method is called Random Network Distillation (RND), and it demonstrated high exploration performance in challenging tasks among the Atari games, achieving state-of-the-art performance at the time.
Eysenbach et al. (2018) proposed a method for learning diverse skills without extrinsic rewards by designing intrinsic rewards based on the mutual information between skills and states \cite{Eysenbach18}.
They demonstrated that these skills are useful for adapting to unseen tasks with minimal exploration.

Active research has been conducted not only on exploration efficiency but also on approaches to improve sample efficiency.
Schwarzer et al. (2023) proposed a method that successfully increased the frequency of experience replay while preventing overfitting \cite{Schwarzer23}.
As a result, their method achieved human-level performance in the Atari games with just two hours of gameplay.
These studies aim to enhance not only performance but also efficient sampling and exploration, even in domains where reinforcement learning has already excelled.

\subsection{Target-oriented exploration}
Target-oriented reinforcement learning methods have been actively studied to improve efficiency in sampling and exploration.
Liu et al. (2022) surveyed problems and solutions related to goal-conditioned reinforcement learning \cite{Liu22}.
The common approaches that they surveyed involves redesigning rewards based on the distance between the current state and the goal state, using the goal state as a condition.
Arumugam et al. (2024) proposed a method that explores solutions under the condition that errors are bounded within a targeted range \cite{Arumugam24}.
However, there is still room for further study because there can be various approaches to define objectives and benchmark methods.

Previously, we proposed Risk-sensitive Satisficing (RS) as a bandit algorithm to enable the agent to perform target-oriented exploration \cite{Takahashi16}.
The RS agent selects the action that maximizes the $\it{I}^{\mathrm{RS}}(a)$ value, defined by the following value function (Eq. (\ref{rs})), from the action space $A$.
\begin{eqnarray}
\it{I}^{\mathrm{RS}}(a) &=& \frac{n(a)}{N} \bigl(E(a) - \aleph \bigr)\qquad \forall a \in A
\label{rs}
\end{eqnarray}
Here, $n(a)$ is the number of times action $a$ has been selected, $N = \sum_{a \in A} n(a)$ is the total number of selections at the time, $E(a)$ is the average reward (action value) obtained by action $a$, and $\aleph$ (aleph) is the aspiration level.
The reliability of the action value $E(a)$ is defined as the agent's selection ratio $n(a) / N$.
RS sets the aspiration level utilizing internal factors (e.g., the necessary energy for survival of an animal) or external factors (e.g., prior task knowledge).

Past studies have demonstrated RS’s superior performance in bandit problems (e.g., $K$-armed bandit \cite{Kamiya22} and contextual bandit \cite{Tsuboya24}) and multi-agent reinforcement learning tasks \cite{Uragami24}.
In particular, it provided performance guarantees in the Bernoulli bandit problem \cite{Tamatsukuri19}.
The regret of standard bandit algorithms grows logarithmically, whereas RS’s regret stops increasing and ultimately keeps upper-bounded by a finite value.
This capability is particularly useful in deep reinforcement learning, where trial and error can be costly.

An early attempt to apply RS to deep reinforcement learning was first made by Satori et al. (2019) \cite{Satori19}.
Following this, Kono et al. (2023) improved their method \cite{Kono23}.
However, these studies are insufficient in terms of verifying RS's exploration capabilities and comparing them with that of standard exploration methods.
Moreover, recent improvements on RS algorithms are not incorporated in these studies.
This study proposes a method that fully integrates past research on RS.
We compared the performance and exploration capability of our method with those of a standard exploration method, RND, using two types of tasks.

\section{Methods}
We propose Regional Stochastic Risk-sensitive Satisficing ($\mathrm{RS}^2$) that extends RS into an exploration policy for deep reinforcement learning.
This exploration policy is applicable to a wide range of off-policy reinforcement learning methods, such as Q-learning, Off-Policy Gradient, and Actor-Critic \cite{Espeholt18}.
In this study, we examined a case where $\mathrm{RS}^2$ was applied to DQN. 
Our method introduces two approaches to generalize RS: (1) Reliability estimation that uses state vectors, and (2) A meta-mechanism for setting the aspiration levels of states.
The rest of this section describes how we applied these approaches to RS.

\subsection{Reliability estimation that uses state vectors}
In tasks with a discrete state space, RS can calculate the reliability $n(a)/N$ of an action $a$ based on the selection ratio of actions.
However, estimating reliability based on simple counts faces difficulties in deep reinforcement learning due to the vast and complex state spaces.
For reliability estimation $\mathrm{RS}^2$ uses, based on the concept of soft clustering, the selection frequency of actions.

We initialized $K$ cluster centroid vectors, $C(a) = \{ c(a,1), c(a,2), \ldots, c(a,K) \}$, for each action class $a$.
Here, $c(a,k)$ is the vector obtained by clustering the states observed when action $a$ was selected, and $C(a)$ is a set of representative state vectors that correspond to action class $a$.
We set the number of centroids for each action class to $K=3$ in this study.
The centroids were initialized by generating samples from a standard normal distribution.
Each vector was then normalized by dividing it by its norm, which means it was scaled to have the norm of 1.

The similarity of two kinds of vectors, the latent representation vector $z_t$ at time $t$ and the centroid vector $c(a,k)$, was calculated using the distance between them.
$\mathrm{RS}^2$ used the hidden layer as $z_t$ directly before the output layer in the feed-forward neural network.
Equation (\ref{distance}) shows the Euclidean distance between vectors, and Equation (\ref{sim}) represents the similarity based on the reciprocal of the distance.
\begin{eqnarray}
\label{distance}
d(a,k) &=& ||z_t - c(a,k)||\\
w(a,k) &=& \frac{1}{d(a,k) + \epsilon}
\label{sim}
\end{eqnarray}
$\epsilon$ is a small coefficient that prevents division by zero.
We designed the similarity $w(a,k)$ to increase as the distance between vectors decreases.

Equation (\ref{mean}) adjusts the selection count $n(a)$ using the average value of the similarity (weights), and Equation (\ref{softmax}) estimates reliability using the softmax function.
\begin{eqnarray}
\label{mean}
\overline{n}(a) &=& \frac{n(a)}{K} \sum_{k\in K} w(a,k)\\
\rho (a) &=&  \mathrm{softmax} \bigl( \overline{\mathit{n}}(\it{a}) \bigr)
\label{softmax}
\end{eqnarray}
Each time action $a$ is selected, $n(a)$ is incremented.
The centroid vectors are updated sequentially using a weighted average with $w(a,k)$ as the weights.
$n(a)$ and $w(a,k)$ for all actions are decreased through multiplication with a forgetting rate of $\gamma = 0.9$.
$\mathrm{RS}^2$ provides additional information for prioritizing actions during exploration.
The agent utilizes the reliability $\rho(a)$ in its action selection process.

\subsection{A meta-mechanism for setting aspiration levels of states}
We introduced a meta-mechanism that dynamically adjusts the aspiration level for each state $\aleph(s)$ based on the difference between an actual return $V_G$ and an ideal return $\aleph_G$.
The agent aims to acquire the return of $\aleph_G$ in the process from the initial state to the terminal state.
On the other hand, the aspiration level for each state $\aleph(s)$ represents the return to be acquired in the course from the current state to the terminal state.
$\aleph(s)$ is less than or equal to $\aleph_G$ because it excludes the rewards obtained from the initial state to the current state, as expressed by $\aleph(s) \leq \aleph_G$.

The ideal behavior of the agent is to explore more at earlier stages of an episode and less exploration at later stages.
The study by Kamiya et al. (2022) demonstrated that the greater RS's aspiration level relative to all action values, the more the agent's exploration becomes random.
When the aspiration level is lower than an action value, the agent stops exploring and keeps selecting the action with the maximum value $\max_a Q(s,a)$ \cite{Kamiya22}.
Based on these properties, we designed $\aleph(s)$ to be closer to $\aleph_G$ at the beginning of an episode and to move closer to $\max_a Q(s,a)$, the maximum action value, as the episode progresses.
Equation (\ref{aleph_s}) defines the aspiration level $\aleph(s)$ for each state.
\begin{eqnarray}
\label{aleph_s}
\aleph(s) &=& \beta \aleph_G + (1-\beta)\max_a Q(s,a)\\
\beta &=& \min \Biggl( \max \biggl( \frac{\aleph_G - V_G}{\aleph_G}, 0 \biggr), 1 \Biggr) \nonumber
\end{eqnarray}
$\beta$ takes the range of [0, 1].
The agent reduces exploration when $\aleph_G \leq V_G$ since $\beta = 0$ and $\aleph(s) = \max_a Q(s,a)$, but when $\aleph_G > V_G$, $\beta$ approaches 1 and the agent increases exploration.
This mechanism automatically adjusts the aspiration level for each state, enabling the designer to simply set a single target, $\aleph_G$.

\subsection{Regional Stochastic Risk-sensitive Satisficing}
$\mathrm{RS}^2$ determines the action to be taken using $\rho(a)$ from Equation (\ref{softmax}) and $\aleph(s)$ from Equation (\ref{aleph_s}).
The $\mathrm{RS}^2$ agent calculates the $\it{I}^{\mathrm{RS}^2}(a)$ value using Equation (\ref{rsrs}) when it has found an action whose value exceeds the aspiration level.
\begin{eqnarray}
\it{I}^{\mathrm{RS}^2}(a) = \rho(a) \bigl( Q(s,a) - \aleph(s) \bigr)\qquad \forall a \in A
\label{rsrs}
\end{eqnarray}
Here, a neural network learns the action value $Q(s,a)$.

The $\mathrm{RS}^2$ agent calculates the Stochastic RS (SRS) value, denoted as $\it{I}^{\mathrm{SRS}}(a)$, using Equation (\ref{srs}) when it has not yet found an action whose value exceeds the aspiration level.
\begin{eqnarray}
\delta(a) &=& \aleph(s) - Q(s,a) \nonumber \\
z &=& \frac{1}{\sum_{a\in A}\delta(a)^{-1}}\nonumber\\
\hat{\rho}(a) &=& \frac{z}{\delta(a)}\nonumber\\
b &=& \max \frac{\rho(a)}{\hat{\rho}(a) + \epsilon}\nonumber\\
\it{I}^{\mathrm{SRS}}(a) &=& b\hat{\rho}(a) - \rho(a) \qquad \forall a \in A
\label{srs}
\end{eqnarray}
The agent generates the exploration policy through using the softmax function on $\it{I}^{\mathrm{RS}^2}(a)$ or $\it{I}^{\mathrm{SRS}}(a)$.

\section{Experiments}
\begin{figure}[t]
\vspace{-4pt}
\begin{center}
\includegraphics[scale=0.12]{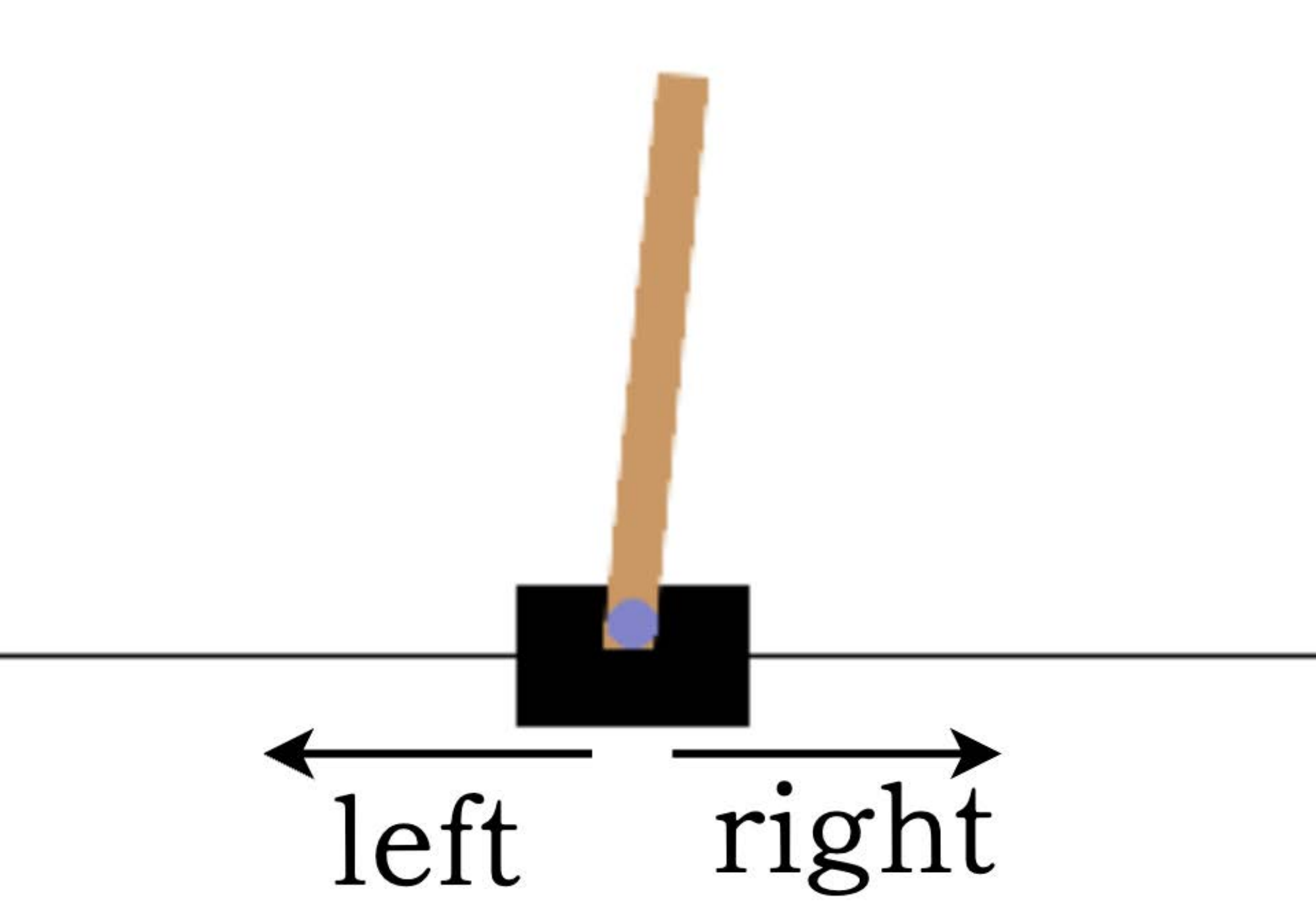}
\end{center}
\vspace{-10pt}
\caption{CartPole-v0 \cite{CartPole} }
\vspace{-4pt}
\label{task:cartpole}
\end{figure}

\begin{figure}[t]
    \begin{tabular}{cc}
      \begin{minipage}[b]{0.51\columnwidth}
        \centering
        \includegraphics[keepaspectratio, scale=0.09]{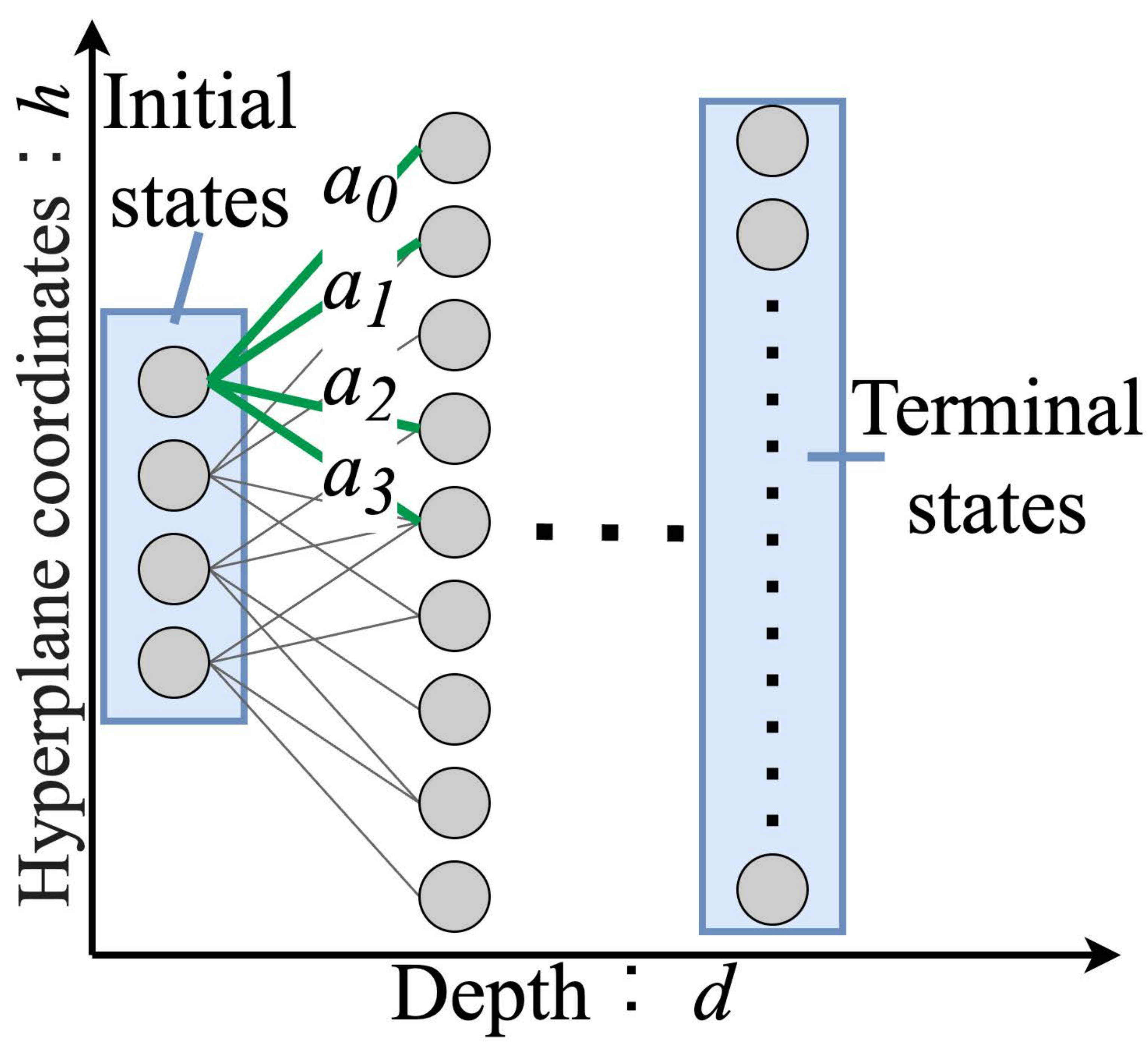}
        \subcaption{Pyramid task}
        \label{pyramid1}
      \end{minipage} &
      \begin{minipage}[b]{0.46\columnwidth}
        \centering
        \includegraphics[keepaspectratio, scale=0.09]{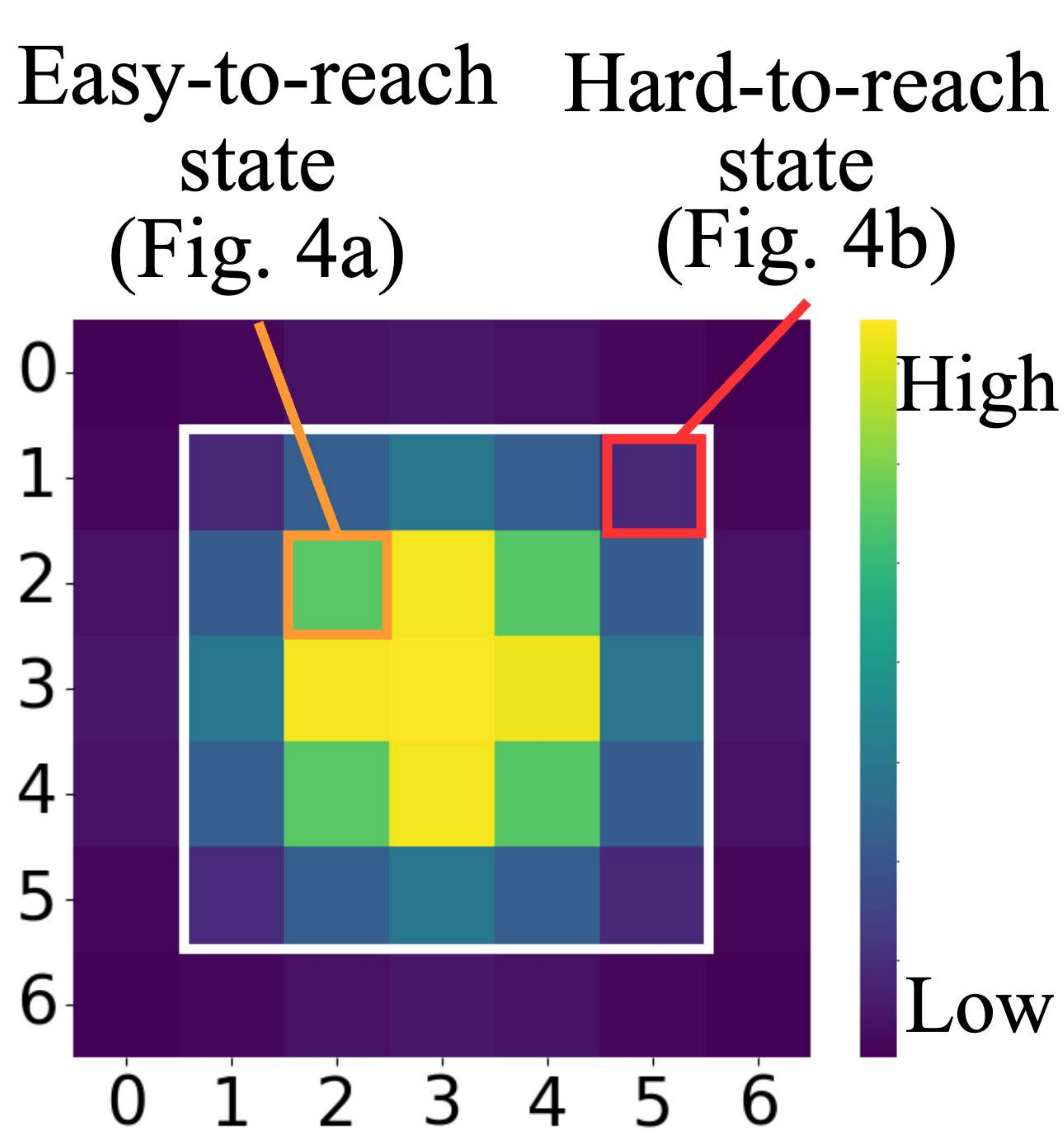}
        \subcaption{Terminal states}
        \label{pyramid2}
      \end{minipage}
    \end{tabular}
    \vspace{-8pt}
    \caption{Overview of Pyramid task \cite{Ikeda22}. (a) An example of Pyramid Task with a depth of 6 and a hyperplane coordinate dimensionality of 2. (b) The frequency of reaching each terminal state by a random agent. The white-bordered area represents the set of states that can be reached from all initial states. In this experiment, rewards were assigned in two patterns: the red-bordered state (hard-to-reach) and the orange-bordered state (easy-to-reach).}
    \label{task:pyramid}
\end{figure}

In this study, we tested the performance of our method using two tasks: a motion control task and a navigation task.
We selected CartPole for the motion control task and Pyramid task for the navigation task.
Figure \ref{task:cartpole} and Figure \ref{task:pyramid} present the overview of each task.
These tasks have distinct characteristics in terms of reward structure and state representation.
In both tasks, we examined whether our method demonstrates stable performance.

\subsection{Motion control task}
CartPole is a motion control task with dense rewards, provided by OpenAI Gym.
The aim of this task is to balance the cart to prevent the pole from falling.
The episode ends either when the pole falls or after 200 steps.
The state is a 4-dimensional vector that includes the positions and velocities of the pole and the cart.
The action consists of two choices: left or right, and the reward is +1 at each step.

The dense rewards in this task make the learning process less likely to stagnate.
On the other hand, appropriate policies are unclear and difficult to learn because the states are represented by continuous values, such as velocities.

\subsection{Navigation task}
Pyramid task is a tree-structured navigation task with sparse rewards.
The aim of this task is to learn a policy to reach the single-reward state at the terminal state.
The agent starts from the initial state and moves toward the terminal state.
The initial state is randomly selected from one of the four states.
We used a Pyramid task with a depth of $d=6$ and a hyperplane coordinate dimensionality of $h=2$.
The number of actions is $4$, the number of states at a depth $d$ is $(1 + d)^h$, and the number of terminal states is $49$.
Each state vector is initialized with random values at the beginning of the simulaion.
This process means representing the states, which are originally continuous, as discrete values.

In this task with sparse rewards, the agent needs to explore efficiently.
On the other hand, the appropriate policies are easy to understand because the state transitions are clear.
A characteristic of Pyramid task is that each of terminal states has different levels of difficulty to reach.
In this experiment, we set the reward for two states, and the two states had different levels of difficulty to reach (Figure \ref{pyramid2}).
We examined the effects of these settings on the agent's performance and exploration capability.

\subsection{Setting}
We used four comparison methods: DQN, DQN with RND (DQN-RND), our method $\mathrm{RS}^2$, and $\mathrm{RS}^2$ with RND ($\mathrm{RS}^2$-RND).
The exploration policy for both DQN and DQN-RND was set to $\epsilon$-greedy.
RND used a neural network with 512 units in each hidden layer and 16 units in the output layer.
The number of the fully connected layers of RND were three layers for CartPole and two layers for Pyramid task.

In CartPole, we used a neural network as the value approximation function for all the methods, a neural network consisting of three fully connected layers with 128 units in each hidden layer.
$\epsilon$ was set to decay exponentially from 1.0 to 0.01 toward the final episode.
The average number of episodes for $V_G$ was 100, and $\aleph_G$ was set to 195.
These values were set based on the definition of success provided in the official leaderboard \cite{CartPole_Leaderboard}.

In Pyramid task, we used a neural network consisting of two fully connected layers and 512 units in each hidden layer.
$\epsilon$ was kept constant at 0.1.
The average number of episodes for $V_G$ was 100, and $\aleph_G$ was set to 1.

The result figures show the average of 100 simulations.
First, we evaluated the performance of our method based on the progress in return obtained by the agent under the estimated policy.
Second, in Pyramid task, we visualized the agent’s visitation frequency of each terminal state and analyzed the agent's exploration tendencies.

\section{Results and Discussion}
In this section, we present the results of each method in CartPole and Pyramid task.

\subsection{Achieved Return}
\begin{figure}[b]
\vspace{-4pt}
\begin{center}
\includegraphics[scale=0.5]{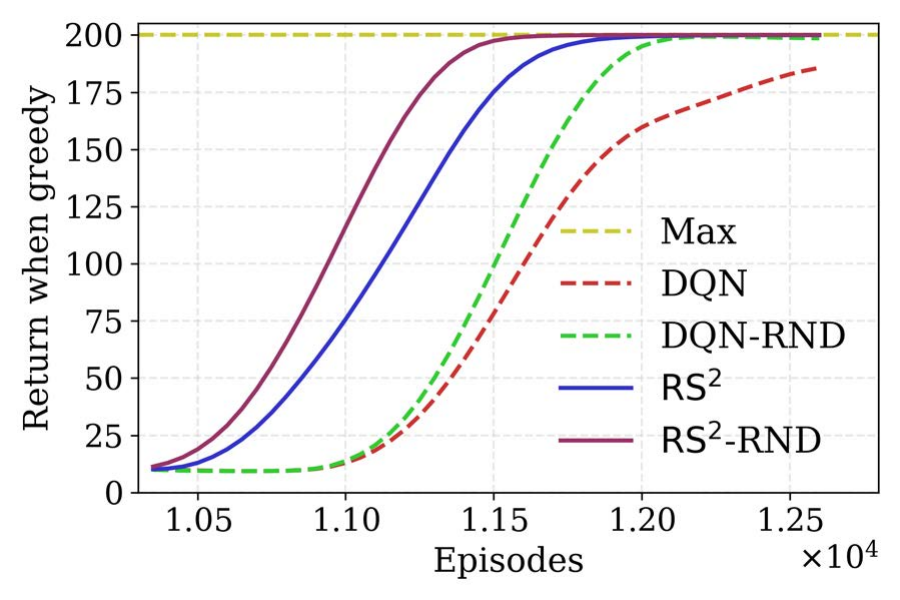}
\end{center}
\vspace{-20pt}
\caption{Return achieved by each method in CartPole}
\vspace{-4pt}
\label{fig:cartpole}
\end{figure}

Figure \ref{fig:cartpole} shows the progress in returns obtained by the estimated policy in CartPole.
$\mathrm{RS}^2$ consistently outperformed the baseline methods, including DQN and DQN-RND.
Notably, our method demonstrated a rapid growth in performance during the earlier episodes.
This rapid growth is a conspicuous characteristic compared to DQN-RND, which used the standard exploration technique, RND.
This characteristic of $\mathrm{RS}^2$ is consistent with the properties of RS demonstrated in past studies.
This result indicates that our method successfully expands the applicability of RS while retaining the properties.

\begin{figure*}[t]
    \begin{tabular}{cc}
      \begin{minipage}[b]{1.0\columnwidth}
        \centering
        \includegraphics[keepaspectratio, scale=0.45]{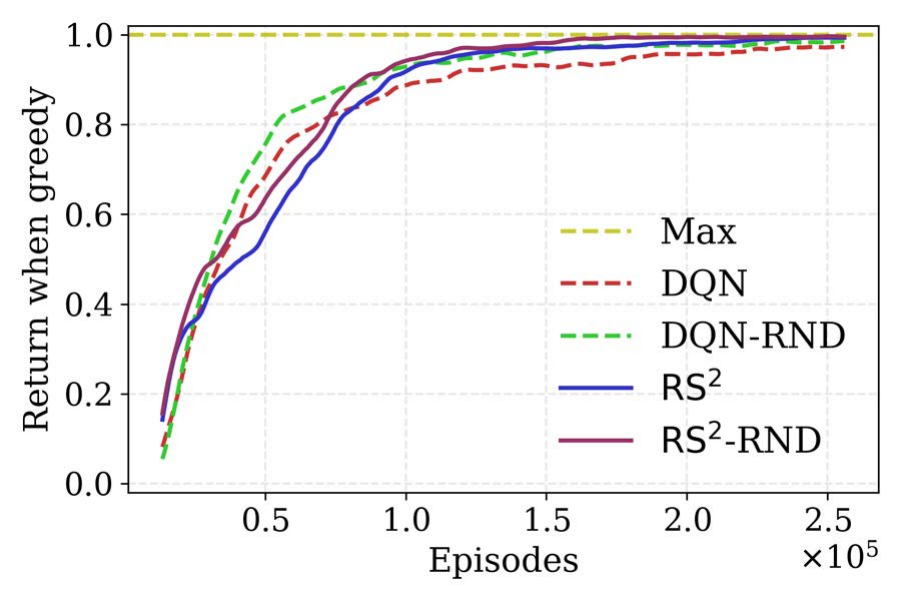}
        \subcaption{An easy-to-reach state with reward}
        \label{returns_622}
      \end{minipage} &
      \begin{minipage}[b]{1.0\columnwidth}
        \centering
        \includegraphics[keepaspectratio, scale=0.45]{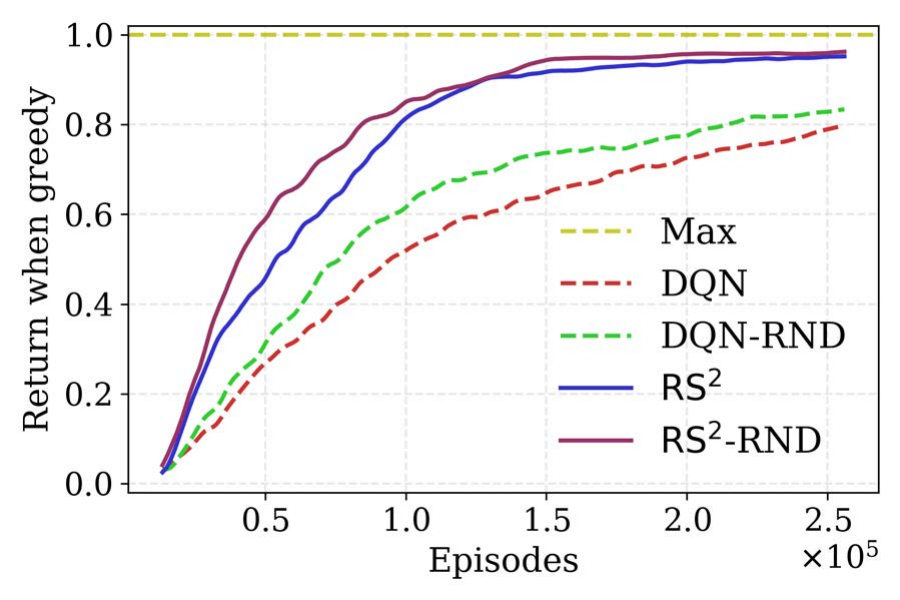}
        \subcaption{A hard-to-reach state with reward}
        \label{returns_615}
      \end{minipage}
    \end{tabular}
     \vspace{-8pt}
     \caption{Return achieved by each algorithm in Pyramid task}
    \label{fig:pyramid}
\end{figure*}

Figure \ref{fig:pyramid} shows the progress in returns obtained by the estimated policy in Pyramid task.
Figure \ref{returns_622} and Figure \ref{returns_615} show that $\mathrm{RS}^2$ consistently achieved high returns regardless of the difference in difficulty to find the reward.
These results showed that our method surpassed performances of both DQN and DQN-RND in situations where finding the reward was more challenging.
RND was an effective method in hard-to-explore environments, but it hardly improved performance in this experiment.
This result was probably caused by the sharp decline in intrinsic rewards, the decline was due to the quick diminishment of the novelty of states in an environment with discrete states and a small-scale state space.

\subsection{Exploration of the Environment}
\begin{figure*}[t]
  \centering
    \begin{tabular}{cccc}
      \begin{minipage}[b]{0.6\columnwidth}
        \centering
        \includegraphics[keepaspectratio, scale=0.085]{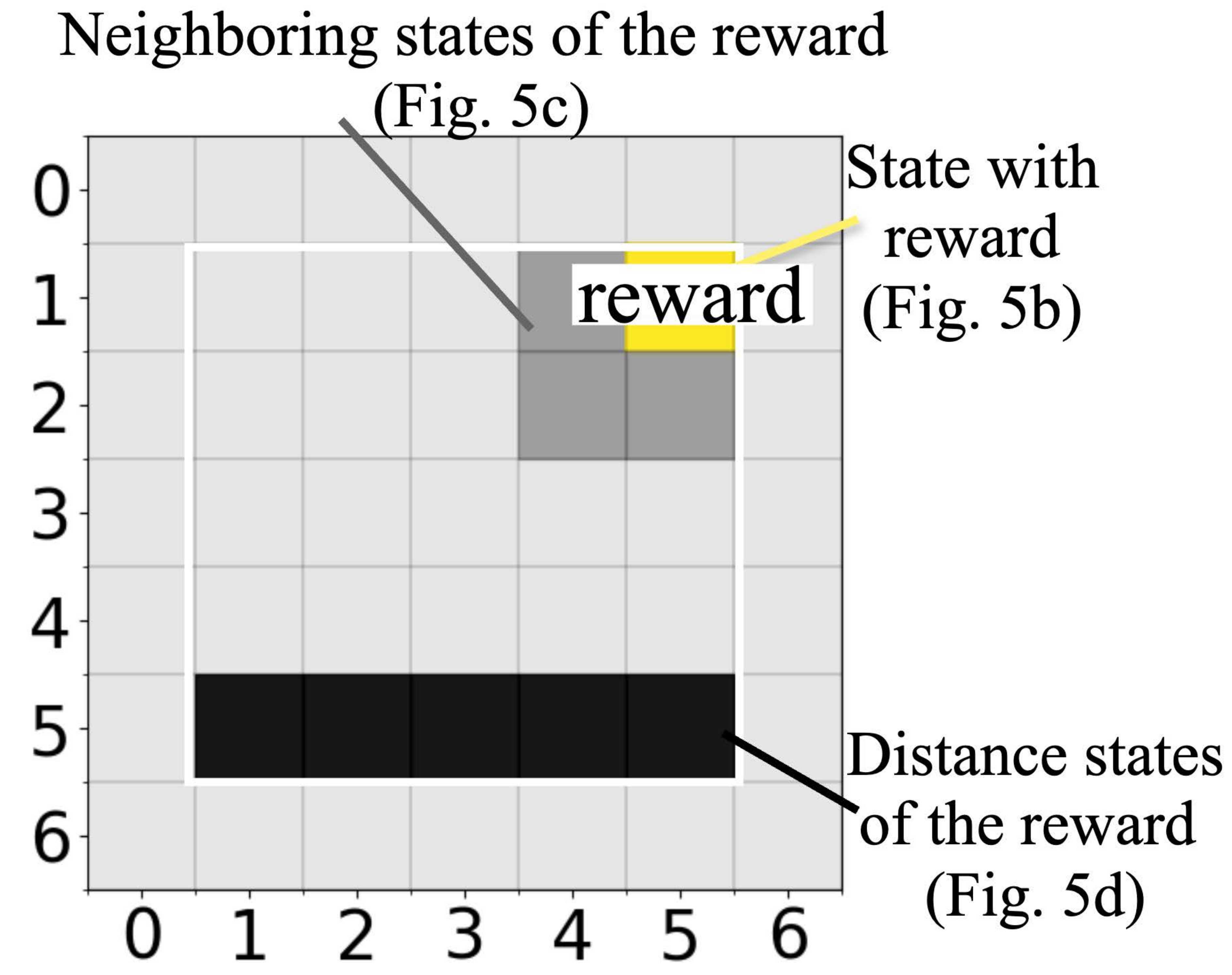}
        \subcaption{Three types of state groups}
        \label{reward}
      \end{minipage} &
      \begin{minipage}[b]{0.46\columnwidth}
        \centering
        \includegraphics[keepaspectratio, scale=0.21]{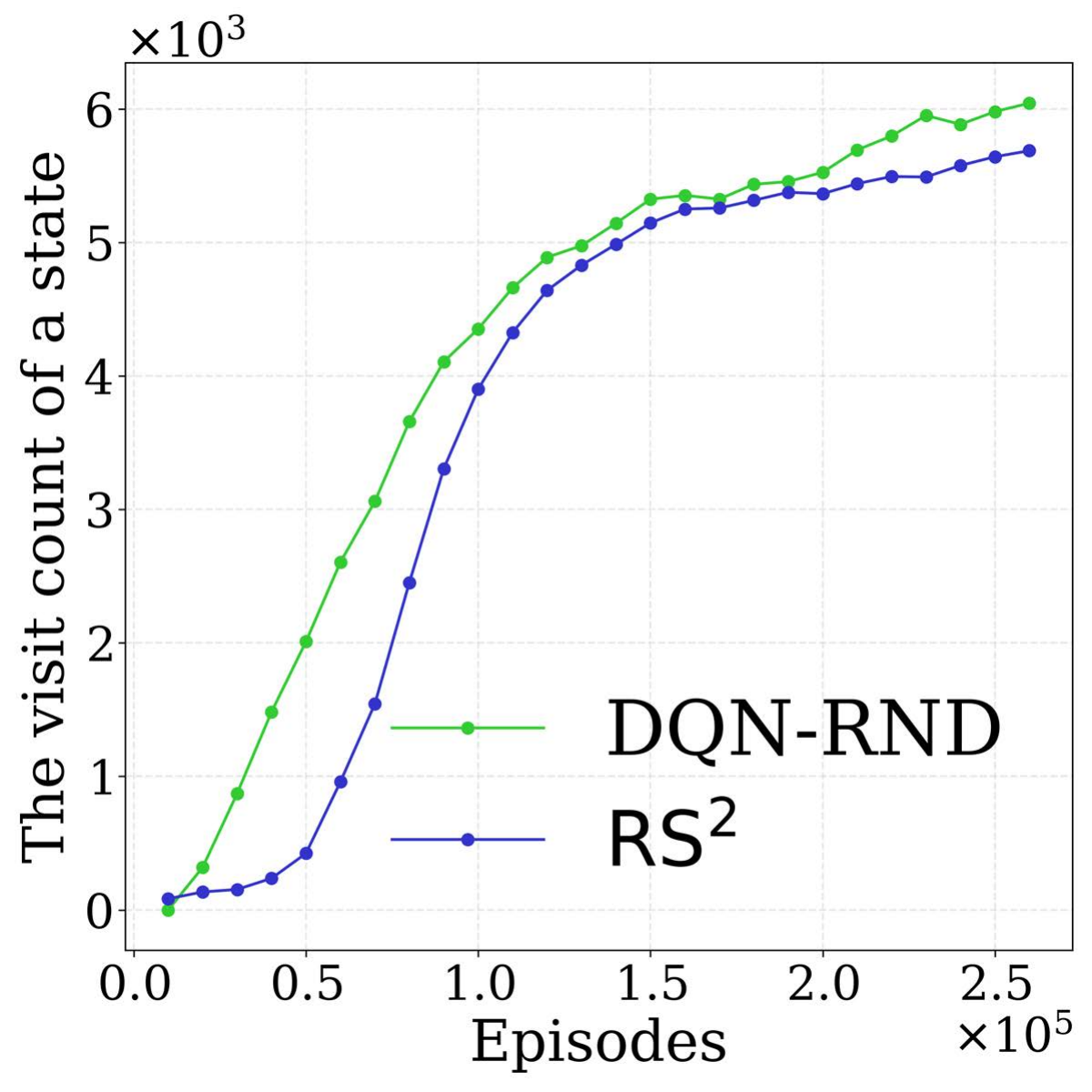}
        \subcaption{The reward state}
        \label{reward_generate}
      \end{minipage} &
      \begin{minipage}[b]{0.46\columnwidth}
        \centering
        \includegraphics[keepaspectratio, scale=0.21]{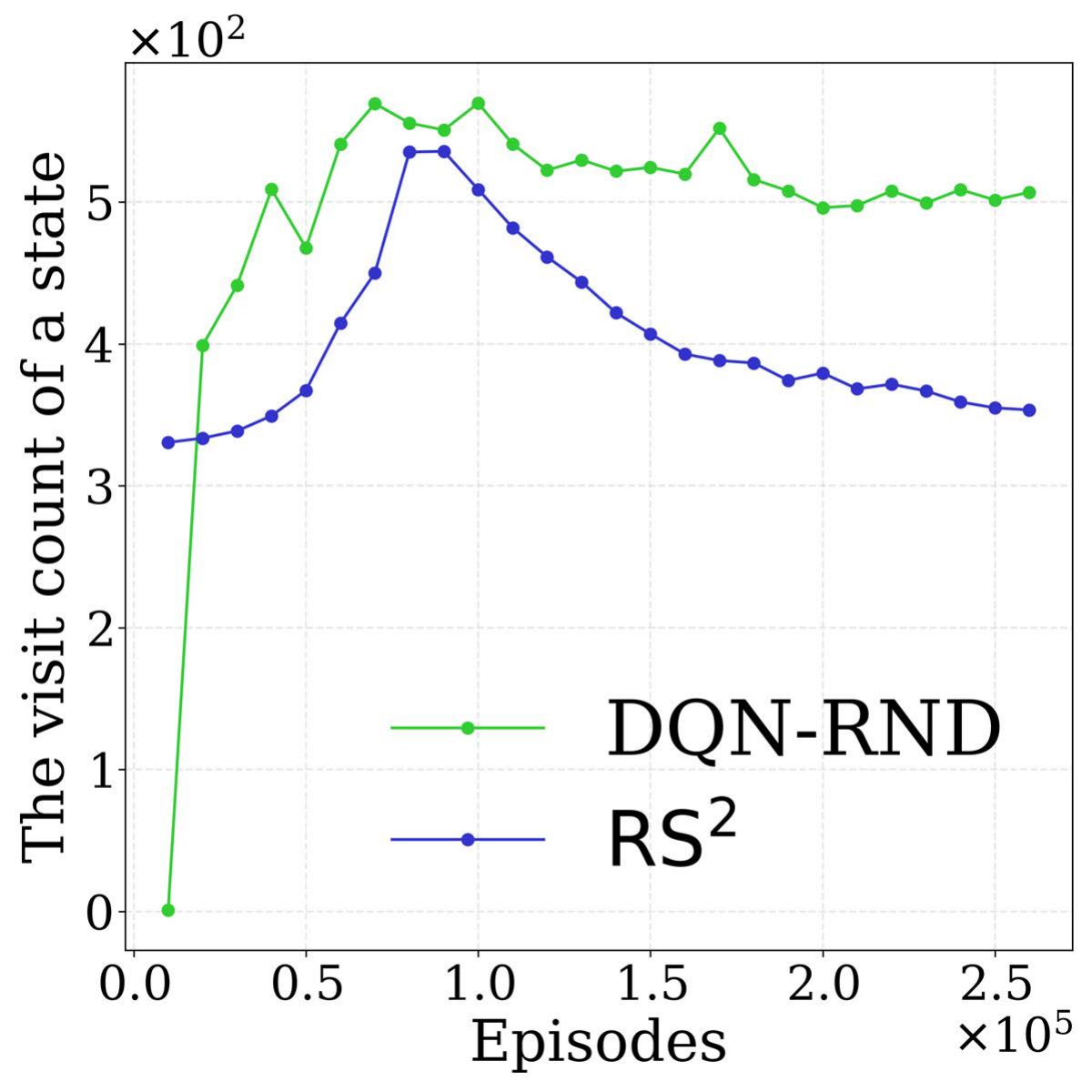}
        \subcaption{The neighboring states}
        \label{others_generate}
      \end{minipage} &
      \begin{minipage}[b]{0.46\columnwidth}
        \centering
        \includegraphics[keepaspectratio, scale=0.21]{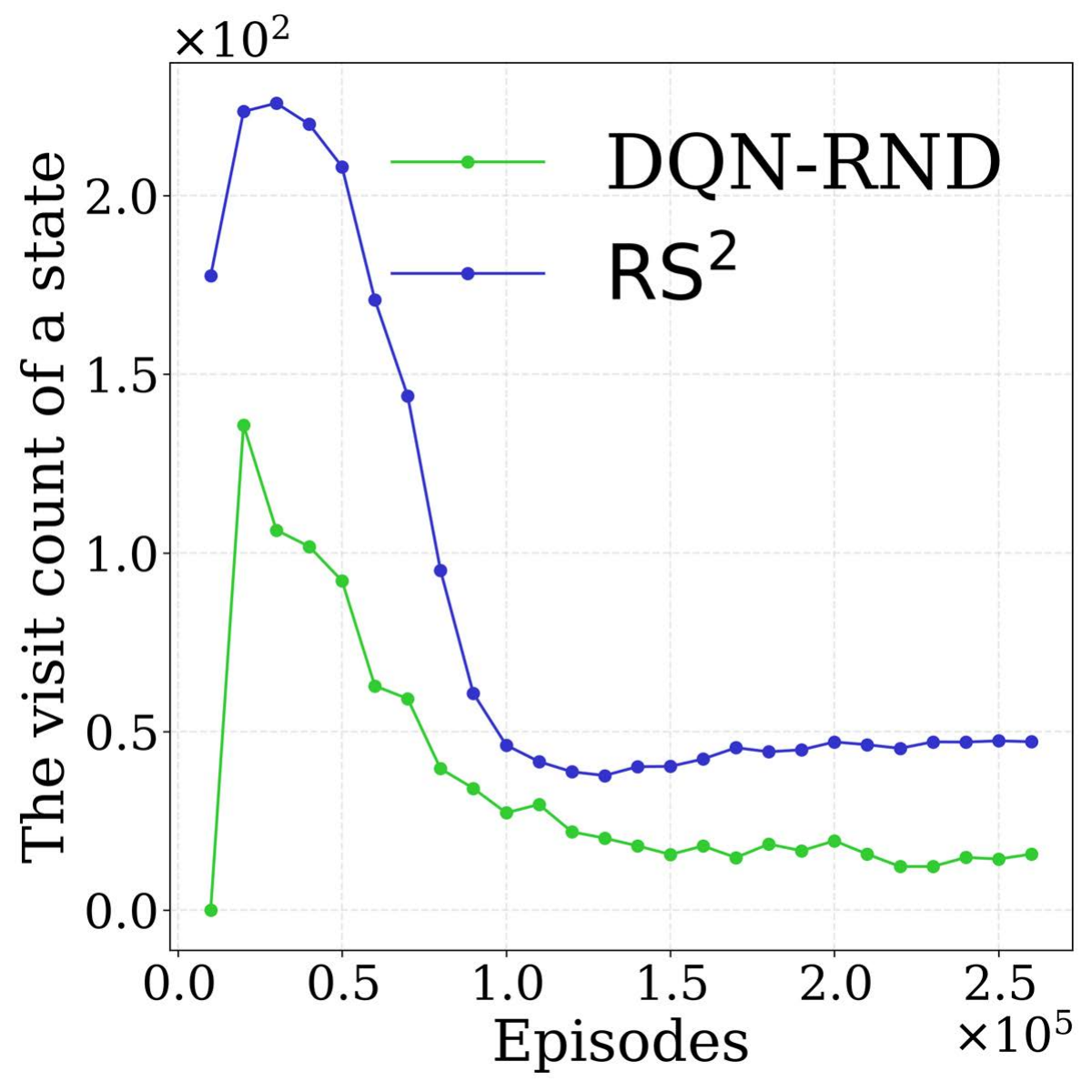}
        \subcaption{The distant states}
        \label{long_generate}
      \end{minipage}
    \end{tabular}
    \vspace{-8pt}
    \caption{Exploration tendencies of each agent. (a) Yellow indicates state with reward. Dark gray indicates groups of states neighboring the reward. Black indicates groups of states distant from the reward. (b), (c), and (d) show the number of times each agent visited the yellow, dark gray, and black states, respectively.}
    \label{generate_policy}
\end{figure*}

We analyzed the characteristics of the agent's exploration using the setup of Figure \ref{returns_615}.
Figure \ref{generate_policy} shows the exploration behavior of DQN-RND agent and $\mathrm{RS}^2$ agent.
$\mathrm{RS}^2$ promoted the agent's learning by expanding the scope of exploration during the initial episodes.
Figure \ref{long_generate} shows that $\mathrm{RS}^2$ actively explored states distant from the reward throughout the first 50,000 episodes, and after that it rapidly narrowed the scope of exploration.
As to the states distant from the rewards, RND demonstrated a lower frequency of exploration compared to $\mathrm{RS}^2$.
During the early stages of learning, the agent needs to expand its exploration scope to find better actions.
Our method promoted exploration during the initial episodes more effectively while maintaining similar exploration tendencies to RND.
As the result, $\mathrm{RS}^2$ successfully achieved high returns from the initial episodes (Figures \ref{fig:cartpole} and \ref{fig:pyramid}).
Moreover, our agent did not completely eliminate exploration of distant states even after its learning had advanced. 
This result suggests that the agent showed adaptability to changes in reward states within non-stationary environments.

$\mathrm{RS}^2$ dynamically adjusted the scope of exploration depending on the agent's learning progress.
From Figure \ref{reward_generate}, we can see that both methods increased the number of explorations that reached the reward state.
Figure \ref{others_generate} shows that our method reduced the number of explorations to the states neighboring the reward after 100,000 episodes.
The reason $\mathrm{RS}^2$ narrowed the scope of exploration was presumably that it achieved an average return of 0.8 by the 100,000th episode.
Even when RND achieved a similar return at the 200,000th episode, it did not narrow the scope of exploration.
These results revealed that our method adjusted the scope of exploration more flexibly based on the progress of learning.

\section{Conclusion}
We proposed $\mathrm{RS}^2$, a deep reinforcement learning method that prioritizes achieving the aspiration level over maximizing expected return.
$\mathrm{RS}^2$ showed the ability to flexibly adjust the exploration scope to achieve the aspiration level.
Commonly used reinforcement learning methods aim to maximize return.
However, these methods have difficulty in explicitly evaluating the current level of target achievement.
For this reason, standard exploration techniques (e.g., RND) have difficulty appropriately adjusting the scope of exploration based on the progress of learning.
On the other hand, our method can easily calculate the level of target achievement based on an expected return and a set aspiration level.
This characteristic enables the agent to flexibly adjust the exploration scope depending on the level of target achievement.

Target-oriented exploration methods are particularly useful in practical applications where quick and efficient achievement of specific aspiration levels is required.
These methods are well-suited for tasks such as power control and inventory management, where the top priority is maintaining supply levels above a threshold.
Furthermore, target-oriented approaches can contribute not only to engineering but also to behavioral modeling in fields such as psychology and cognitive science, and provide new perspectives in understanding human and animal behavior.


\begin{thebibliography}{10}
\providecommand{\url}[1]{#1}
\csname url@samestyle\endcsname
\providecommand{\newblock}{\relax}
\providecommand{\bibinfo}[2]{#2}
\providecommand{\BIBentrySTDinterwordspacing}{\spaceskip=0pt\relax}
\providecommand{\BIBentryALTinterwordstretchfactor}{4}
\providecommand{\BIBentryALTinterwordspacing}{\spaceskip=\fontdimen2\font plus
\BIBentryALTinterwordstretchfactor\fontdimen3\font minus \fontdimen4\font\relax}
\providecommand{\BIBforeignlanguage}[2]{{%
\expandafter\ifx\csname l@#1\endcsname\relax
\typeout{** WARNING: IEEEtran.bst: No hyphenation pattern has been}%
\typeout{** loaded for the language `#1'. Using the pattern for}%
\typeout{** the default language instead.}%
\else
\language=\csname l@#1\endcsname
\fi
#2}}
\providecommand{\BIBdecl}{\relax}
\BIBdecl

\bibitem{Silver17}
D.~Silver, J.~Schrittwieser, K.~Simonyan \emph{et~al.}, ``Mastering the game of {{Go}} without human knowledge,'' \emph{Nature}, vol. 550, no. 7676, pp. 354--359, 2017.

\bibitem{Kool19}
W.~Kool, H.~van Hoof, and M.~Welling, ``Attention, {{Learn}} to {{Solve Routing Problems}}!'' in \emph{Proc. of {{ICLR}}}, 2019.

\bibitem{Degrave22}
J.~Degrave, F.~Felici, J.~Buchli \emph{et~al.}, ``Magnetic control of tokamak plasmas through deep reinforcement learning,'' \emph{Nature}, vol. 602, no. 7897, pp. 414--419, 2022.

\bibitem{Mnih15}
V.~Mnih, K.~Kavukcuoglu, D.~Silver \emph{et~al.}, ``Human-level control through deep reinforcement learning,'' \emph{Nature}, vol. 518, no. 7540, pp. 529--533, 2015.

\bibitem{Schwarzer23}
M.~Schwarzer, J.~{Obando-Ceron}, A.~Courville \emph{et~al.}, ``Bigger, {{Better}}, {{Faster}}: {{Human-level Atari}} with human-level efficiency,'' in \emph{Proc. of {{ICML}}}, 2023.

\bibitem{Takahashi16}
T.~Takahashi, Y.~Kohno, and D.~Uragami, ``{Cognitive Satisficing: Bounded Rationality in Reinforcement Learning},'' \emph{Transactions of the Japanese Society for Artificial Intelligence}, vol.~31, no.~6, pp. AI30--M\_1--11, 2016, (in Japanese).

\bibitem{Tamatsukuri19}
A.~Tamatsukuri and T.~Takahashi, ``Guaranteed satisficing and finite regret: {{Analysis}} of a cognitive satisficing value function,'' \emph{Biosystems}, vol. 180, pp. 46--53, 2019.

\bibitem{Burda18}
Y.~Burda, H.~Edwards, A.~Storkey, and O.~Klimov, ``Exploration by {{Random Network Distillation}},'' in \emph{Proc. of {{ICML}}}, 2018.

\bibitem{Eysenbach18}
B.~Eysenbach, A.~Gupta, J.~Ibarz, and S.~Levine, ``Diversity is {{All You Need}}: {{Learning Skills}} without a {{Reward Function}},'' in \emph{Proc. of {{ICLR}}}, 2019.

\bibitem{Liu22}
M.~Liu, M.~Zhu, and W.~Zhang, ``Goal-{{Conditioned Reinforcement Learning}}: {{Problems}} and {{Solutions}},'' in \emph{Proc. of {{IJCAI}}}, 2022.

\bibitem{Arumugam24}
D.~Arumugam, S.~Kumar, R.~Gummadi, and B.~Van~Roy, ``Satisficing {{Exploration}} for {{Deep Reinforcement Learning}},'' in \emph{Proc. of RLC Finding the Frame Workshop}, 2024.

\bibitem{Kamiya22}
T.~Kamiya and T.~Takahashi, ``Softsatisficing: {{Risk-sensitive}} softmax action selection,'' \emph{Biosystems}, vol. 213, no. 104633, 2022.

\bibitem{Tsuboya24}
A.~Tsuboya, Y.~Kono, and T.~Takahashi, ``{A Sequential Decision-Making Model in Contextual Foraging Behavior},'' \emph{Journal of Japan Society for Fuzzy Theory and Intelligent Informatics}, vol.~36, no.~1, pp. 589--600, 2024, (in Japanese).

\bibitem{Uragami24}
D.~Uragami, N.~Sonota, and T.~Takahashi, ``Social satisficing: {{Multi-agent}} reinforcement learning with satisficing agents,'' \emph{BioSystems}, vol. 243, no. 105276, 2024.

\bibitem{Satori19}
K.~Satori, Y.~Yoshida, T.~Kamiya, and T.~Takahashi, ``{Toward Deep Satisficing Reinforcement Learning},'' in \emph{Proc. of {JSAI}}, 2019, (in Japanese).

\bibitem{Kono23}
Y.~Kono, J.~Kume, R.~Ikeda, and T.~Takahashi, ``{Target-oriented Exploration in Deep Reinforcement Learning},'' in \emph{Proc. of {JSAI}}, 2023, (in Japanese).

\bibitem{Espeholt18}
L.~Espeholt, H.~Soyer, R.~Munos \emph{et~al.}, ``Impala: Scalable distributed deep-rl with importance weighted actor-learner architectures,'' in \emph{Proc. of {{ICML}}}, 2018.

\bibitem{CartPole}
G.~Brockman, V.~Cheung, L.~Pettersson \emph{et~al.}, ``Openai gym,'' \emph{arXiv preprint arXiv:1606.01540}, 2016.

\bibitem{Ikeda22}
R.~Ikeda, A.~Minami, Y.~Kono, and T.~Takahashi, ``Developing a scalable and simple verification task of deep reinforcement learning,'' in \emph{Proc. of {JSAI}}, 2022, (in Japanese).

\bibitem{CartPole_Leaderboard}
{OpenAI}, ``{OpenAI Gym Leaderboard: CartPole-v0},'' \url{https://github.com/openai/gym/wiki/Leaderboard#cartpole-v0}, accessed: 2024-11-05.

\end{thebibliography}

\end{document}